\documentclass[conference]{IEEEtran}
\IEEEoverridecommandlockouts
\usepackage{cite}
\usepackage{amsmath,amssymb,amsfonts}
\usepackage{amsmath, amssymb}
\usepackage{algorithmic}
\usepackage{graphicx}
\usepackage{textcomp}
\usepackage{xcolor}
\usepackage{enumitem}
\usepackage{multirow}
\usepackage{hhline}
\usepackage{array}
\usepackage{multirow}
\usepackage{booktabs}
\usepackage{makecell}   

\def\BibTeX{{\rm B\kern-.05em{\sc i\kern-.025em b}\kern-.08em
    T\kern-.1667em\lower.7ex\hbox{E}\kern-.125emX}}
\begin{document}

\title{Neural Federated Learning for Livestock \\Growth Prediction 


\thanks{This research was supported by funding from Food Agility CRC Ltd, Australia, funded under the Commonwealth Government CRC
Program. The CRC Program supports industry-led collaborations between industry, researchers and the community. This research
was also partially supported by funding from Meat and Livestock Australia (MLA). The funders had no role in the study design, data
collection, analysis, interpretation of results, writing of the manuscript, or decision to submit for publication.}
}

\author{
\IEEEauthorblockN{Shoujin Wang}
\IEEEauthorblockA{
\textit{University of Technology Sydney}\\
Sydney, Australia \\
shoujin.wang@uts.edu.au
}
\and
\IEEEauthorblockN{Mingze Ni}
\IEEEauthorblockA{
\textit{University of Technology Sydney}\\
Sydney, Australia \\
mingze.ni@uts.edu.au
}
\and
\IEEEauthorblockN{Wei Liu}
\IEEEauthorblockA{
\textit{University of Technology Sydney}\\
Sydney, Australia \\
wei.liu@uts.edu.au
}

\and
\IEEEauthorblockN{Victor W. Chu}
\IEEEauthorblockA{
\textit{University of Technology Sydney}\\
Sydney, Australia \\
wingyan.chu@uts.edu.au
}

\and
\IEEEauthorblockN{Bryan Zheng}
\IEEEauthorblockA{
\textit{University of Technology Sydney}\\
Sydney, Australia \\
boyuan.zheng@uts.edu.au
}
\and
\IEEEauthorblockN{Ayush Kanwal}
\IEEEauthorblockA{
\textit{AgriWebb}\\
Sydney, Australia \\
ayush.kanwal@agriwebb.com
}
\and
\IEEEauthorblockN{Roy Jing Yang}
\IEEEauthorblockA{
\textit{Queensland University of Technology}\\
Brisbane, Australia \\
 roy.j.yang@qut.edu.au
}
\and
\IEEEauthorblockN{Kenneth Sabir}
\IEEEauthorblockA{
\textit{AgriWebb}\\
Sydney, Australia \\
kenny.sabir@agriwebb.com
}
\and
\IEEEauthorblockN{Fang Chen}
\IEEEauthorblockA{
\textit{University of Technology Sydney}\\
Sydney, Australia \\
fang.chen@uts.edu.au
}
}

\maketitle

\begin{abstract}
Livestock growth prediction is essential for optimising farm management and improving the efficiency and sustainability of livestock production, yet it remains underexplored due to limited large-scale datasets and privacy concerns surrounding farm-level data. Existing biophysical models rely on fixed formulations, while most machine learning approaches are trained on small, isolated datasets, limiting their robustness and generalisability. To address these challenges, we propose LivestockFL, the first federated learning framework specifically designed for livestock growth prediction. LivestockFL enables collaborative model training across distributed farms without sharing raw data, thereby preserving data privacy while alleviating data sparsity, particularly for farms with limited historical records. The framework employs a neural architecture based on a Gated Recurrent Unit combined with a multilayer perceptron to model temporal growth patterns from historical weight records and auxiliary features. We further introduce LivestockPFL, a novel personalised federated learning framework that extends the above federated learning framework with a personalized prediction head trained on each farm’s local data, producing farm-specific predictors. Experiments on a real-world dataset demonstrate the effectiveness and practicality of the proposed approaches.

\end{abstract}


\section{Introduction}

Livestock growth modeling and prediction play a critical role in optimising farm management, improving production efficiency, supporting sustainable food systems, and ensuring the security and resilience of the global food supply chain. Accurate and reliable growth predictions enable data-driven and more informed decisions on feeding and grazing strategies, stocking rates, and market timing, thereby enhancing productivity while reducing economic and environmental risks. This importance is particularly pronounced in agriculture-dominated countries like Australia where livestock production operates across vast and climatically diverse regions, and is increasingly challenged by climate variability and resource constraints~\cite{henry2012livestock}.

Although livestock growth prediction is highly relevant to both academic research and industry practice, publicly available studies in this domain remain scarce. The limited body of existing literature indicates that livestock growth modeling is an underexplored research area, highlighting a clear need for further systematic investigation. Existing studies can generally be classified into two main categories based on the technical approaches employed: \textbf{(1) biophysical approaches} and \textbf{(2) machine learning approaches.} Biophysical approaches primarily rely on mathematical models (e.g., GrazFeed~\cite{GrazFeedsoftware}, GRAZPLAN~\cite{donnelly1997grazplan}) to simulate interactions among animals, pastures, soil, and climate in order to forecast livestock weight gain, body composition, and resource requirements. While these models embed strong domain knowledge and predefined mathematical formulations, they often have the difficulty in parameterising at scale, as it requires precise inputs on feed quality and quantity that farmers often lack, especially in complex real-world settings where data across farms and regions are heterogeneous, noisy, and sparse. 
In contrast, machine learning approaches train predictive models directly on collected livestock growth data and subsequently use these models to forecast growth outcomes~\cite{kanwal2025livestock}. As these approaches are data-driven and tailored to specific datasets from individual farms or regions~\cite{tuyishime2025data}, they are generally more flexible and better able to generalize to diverse real-world scenarios.


In recent years, machine learning approaches have attracted increasing attention due to their strong capability to model complex data and deliver accurate and reliable predictions, and have been widely applied in the livestock sector, including tasks such as weight estimation and precision farming~\cite{garcia2020systematic}. However, studies specifically focused on machine learning–based livestock growth prediction remain limited. According to our investigation, although many related studies use the term “weight prediction,” they primarily apply machine learning models to estimate livestock live weight from body indicators such as height and ribeye area~\cite{garro2024enhancing}. These approaches do not predict livestock growth by forecasting future weights over time. Moreover, most existing studies rely on one or more small-scale datasets, typically containing fewer than 1,000 livestock samples, which substantially limits model generalization and reduces their practicality in real-world settings where data are often large-scale, complex, and highly heterogeneous. This reveals a major barrier to advancing machine learning–based livestock growth prediction: the lack of dedicated large datasets that intensively and continuously record livestock growth. In practical farming operations, frequent livestock weighing and the associated data collection and storage are highly time-consuming and labor-intensive, further constraining data availability.






Another factor hindering research on machine learning–based livestock growth prediction is the privacy and security of farm management data. In real-world settings, farmers are often reluctant to share operational and production data, as such data may contain sensitive information related to business practices and is critical for preserving farm-level competitive advantage. Collecting and centralizing data from multiple farms on an external platform can therefore raise significant concerns regarding data privacy, security, and the risk of information leakage. Federated learning, first proposed in 2017 ~\cite{zhao2018federated}, provides a natural and effective solution to these challenges by enabling machine learning models to be trained locally on each client’s data without transferring raw data to any external or centralized storage, thereby preserving data privacy and security. Despite its suitability, to the best of our knowledge, federated learning has not yet been explored for livestock growth prediction. The only closely related work in the literature applies federated learning to crop yield prediction~\cite{manoj2022federated}.

To this end, in order to address the aforementioned gaps in livestock growth prediction, this paper proposes a novel neural federated learning framework specifically designed for livestock growth prediction, termed LivestockFL. In LivestockFL, a Gated Recurrent Unit (GRU) model combined with a multilayer perceptron (MLP) is first adopted as the base prediction model to forecast future livestock weights using historical weight records and auxiliary features (e.g., livestock attributes and location) as inputs. Building upon this model, a federated learning scheme is developed to enable joint training across farms while preserving data locality. Specifically, each farm is treated as an independent client that retains its data in local storage without uploading potentially sensitive livestock production data to a centralized platform. During training, model updates are computed locally on each client using farm-specific data and then transmitted to a central server, where they are aggregated to update the global model. The updated global model parameters are subsequently distributed back to clients for the next training round. Through this iterative process, the global model is collaboratively trained using livestock growth data distributed across multiple farms.


To further enhance the prediction performance of the global model, we propose a personalized extension of the aforementioned federated framework for livestock growth prediction, termed LivestockPFL. LivestockPFL first trains a general federated livestock growth prediction model using LivestockFL, and then fine-tunes the global model on each farm’s local data. This personalization process enables the model to better capture farm-specific characteristics, thereby improving prediction accuracy and reliability for individual farms.




The main contributions of this work are summarized as follows:
\begin{itemize}[leftmargin=*]
\item We propose the first federated learning framework for livestock growth prediction, which enables collaborative model training across distributed farm-level edge devices without centralising sensitive or private data. By allowing multiple farms to jointly train a shared model while preserving data privacy, the proposed framework effectively mitigates data sparsity and data scarcity issues, particularly for small farms with limited historical records. This provides a practical and scalable solution for deploying machine learning–based livestock growth prediction in real-world agricultural settings.

\item Building upon this framework, we further introduce a personalised federated learning approach that extends the standard federated learning framework with a personalized prediction head which is trained on each farm’s local production data. This yields farm-specific growth prediction models that better capture local conditions and management practices, thereby improving prediction accuracy and reliability for individual farms.

\item We conduct extensive experiments on a large-scale real-world livestock production dataset, and the results consistently demonstrate the effectiveness and robustness of the proposed federated and personalised learning frameworks.
\end{itemize}

\section{Related Work}

\subsection{Biophysical Livestock Modelling}
Biophysical models have been used to simulate livestock growth by explicitly modelling feed intake, digestion and metabolism, and their interactions with pasture, soil, and climate. In Australia, the \textbf{SCA Feeding Standards} provide widely used nutrition-requirement relationships \cite{SCA}, while tools such as \textbf{GrazFeed} operationalise these feeding-standard concepts to predict voluntary intake, nutrient supply, and animal production responses under varying diets and pasture conditions \cite{GrazFeed}. At the system level, \textbf{GRAZPLAN} integrates animal, pasture, and soil components to forecast liveweight change under alternative management and seasonal scenarios \cite{donnelly1997grazplan}, and \textbf{SGS}  has been used as a whole-farm/grazing-system simulator combining pasture growth, feed supply, and animal demand to analyse livestock performance across variable environments \cite{sgs, sgs2}. Despite their interpretability and suitability for scenario analysis, these models are rarely end-to-end and often require multiple intermediate estimates (e.g., metabolism and intake) and calibrated parameters, which can cause errors to propagate through the modelling pipeline. In addition, many underlying processes are difficult to observe at scale, leading to systematic bias when key inputs are misspecified or poorly measured. Their transferability across regions and climates can also be limited because core assumptions and parameter settings may not remain valid under geographical or environmental change.

\subsection{Machine Learning for Livestock Growth}
Machine learning (ML) methods provide a data-driven alternative for predicting livestock weight and supporting precision livestock management, and recent reviews report growing adoption of ML for monitoring, decision support, and productivity-related prediction \cite{garcia2020systematic}. Many ML studies focus on current liveweight estimation by fitting regression models (e.g., RF/SVR/NN) to manually collected morphometric measurements such as body length, heart girth, and wither/hip height \cite{ml1,ml2}. Others replace manual measurements with computer vision(CV) pipelines that learn body-shape features from images (e.g., RGB-D) and directly regress to weight, which can reduce handling but depends on controlled image capture and can be sensitive to pose, occlusion, and camera setup \cite{mlcv1}.
Despite this progress, there remains a notable gap in ML models that perform longitudinal growth forecasting from historical weight trajectories, where the learning problem requires modelling temporal dynamics under irregular sequence, missing records, and measurement noise. Such forecasting models are practically important because they enable proactive management by predicting future weights for planning feeding, drafting, and marketing decisions, rather than only estimating an animal’s current weight at observation time \cite{garcia2020systematic}.

\subsection{Federated Learning in Agriculture}
Federated learning (FL) enables multiple clients to collaboratively train a shared model without exchanging raw data by iteratively sending local model updates to a coordinating server \cite{mcmahan2017communication}. In agriculture, this privacy-preserving paradigm is attractive for farm management because production data are often sensitive and constrained by ownership and commercial considerations. Prior work has mainly studied FL from a \textbf{crop-focused} perspective, for example demonstrating benefits for crop-yield prediction using distributed data while avoiding direct data sharing \cite{manoj2022federated,garro2025systematic}. In contrast, FL remains underexplored for \textbf{livestock weight} and we are not aware of studies that develop and evaluate FL specifically for livestock growth forecasting from longitudinal weight histories. This gap is practically important because multi-farm livestock data are typically heterogeneous across farms in genetics, management, and measurement practices, which can challenge standard FL aggregation and motivate personalized or heterogeneity-aware training strategies \cite{mcmahan2017communication}.

\section{Problem Formalization}
In our work, we leverage a diverse set of data types from multiple sources to predict livestock growth. Specifically, for each animal, we incorporate both dynamic data—namely, sequences of historical weight records—and static features, including animal attributes (e.g., sex and breed) and location-related information (e.g., state, NRM region, and farm). Let $i$ index livestock individuals and $t$ index observation times.
Each individual $i$ has an irregular time series of length $T_i$. $\mathbf{x}_{i,t} \in \mathbb{R}^{d_n}, \quad
\mathbf{m}_{i,t} \in \mathbb{R}^{d_m}, \quad
\mathbf{c}_i = (c_{i,1}, \dots, c_{i,K}),$ represent the weight record related numerical features at each time step, masking vector to indicate if there exists an observed weight record or not at the current time step, and statical categorical features of the $i^{th}$ animal respectively. $H$ \text{denotes the prediction horizon.}

Mathematically, given an individual animal $i$'s historical weight records till to the current time point $T_i$, together with the animal's static features, we aim to build a prediction model $M$ to predict the future weights $y_{i,h}$ of the animal $(h = 1, \dots, H)$. $H$ \text{denotes the prediction horizon.} Here each time step in the weight record sequence corresponds to one-month period in our work.



\section{Methodology}
\begin{figure*}[htbp]
\centering{\includegraphics[width=0.8\textwidth]{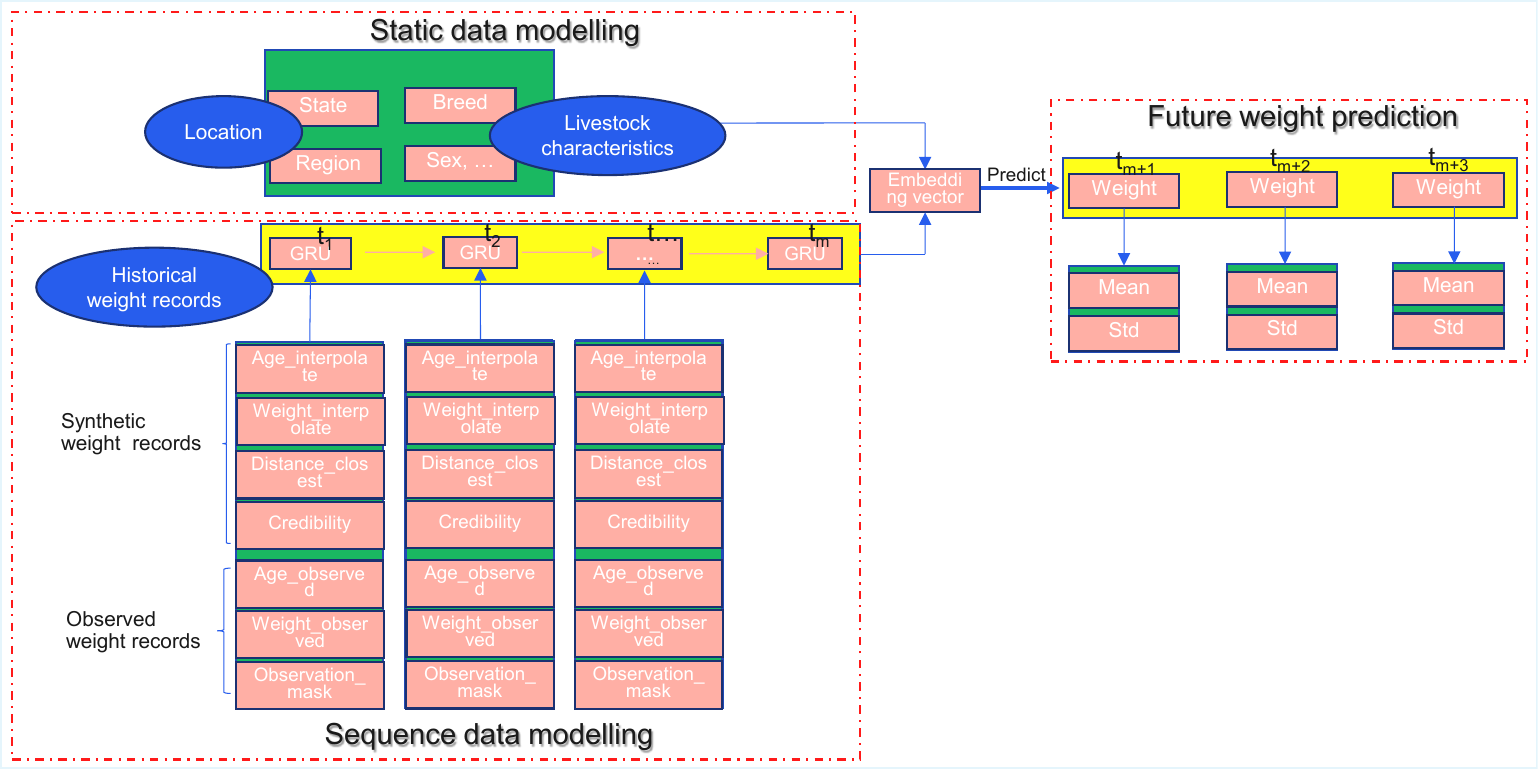}}
\caption{Architecture of our proposed livestock weight prediction model.}
\label{prediction_model}
\end{figure*}

In this section, we will illustrate our proposed method for neural federated learning for livestock growth prediction. Specifically, first, we will describe the base machine learning prediction model, which is built on the top of the commonly utilized time series model Gated Recurrent Unit (GRU) combined with the multiple perception layer (MLP), as shown in Fig \ref{prediction_model}. The GRU is employed to model the time series weight record data while the MLP is to model the static feature data as well as to combine these two parts together in the modeling process. Then, we will describe the federated learning architecture built on the top of the above prediction model, which actually converts the commonly used centralized training scheme to the federated training scheme for data privacy and security. Finally, we will describe our proposed personalized federated learning architecture to further enhance the prior federated learning architecture to a farm-level personalized edition so that the model can be more sensitive and reliable to each farm's specific data.    

\subsection{GRU-based Livestock Growth Prediction Model}

\textbf{Static data embedding.}\label{Encoder} For the static feature data, all of them are categorical ones, and we employ a typical categorical feature embedding layer to transfer each categorical value into a numerical vector, so that it can be well fed into the machine learning model. Specifically, we first embed each categorical value, e.g., $c_{i,l}$ from the $l^{th}$ categorical feature, and then concatenate all categorical feature embeddings together to form a unified categorical embedding vector $\mathbf{e}_{i,1}^{(c)}$, 
\begin{equation}
\mathbf{e}_{i,l}^{(c)}
=
\mathrm{Embedding}_l(c_{i,l})
\in \mathbb{R}^{d_e},
\end{equation}
\begin{equation}
\mathbf{e}_i^{(c)}
=
\left[
\mathbf{e}_{i,1}^{(c)} ; \dots ; \mathbf{e}_{i,L}^{(c)}
\right]
\in \mathbb{R}^{L*d_e}.
\end{equation}

\textbf{Sequence data embedding and modeling.} Once we got the numerical feature input $\mathbf{x}_{i,t}$ for time step $t$ of the $i^{th}$ animal, we first embed it into a latent vector $\mathbf{e}_{i,t}$ and then combined it with the masking vector $\mathbf{m}_{i,t}$ to construct the input for GRU model via the following equations respectively: 
\begin{equation}
\mathbf{e}_{i,t}
= \phi_n(\mathbf{x}_{i,t})
= \mathbf{W}_n \mathbf{x}_{i,t} + \mathbf{b}_n,
\quad
\mathbf{e}_{i,t} \in \mathbb{R}^{d_e}.
\end{equation} 
\begin{equation}
\mathbf{z}_{i,t}
=
\left[
\mathbf{e}_{i,t} \, ; \, \mathbf{m}_{i,t}
\right]
\in \mathbb{R}^{d_e + d_m},
\end{equation}

Then, we take the constructed input vector $\mathbf{z}_{i,t}$ at each time step into the GRU model for modeling the temporal correlations over the historical weight records of each animal and obtain the final hidden state $\mathbf{h}_i$ from the last time step $T_i$ of GRU model as the input of the prediction layer. Specifically,   
\begin{equation}
\mathbf{h}_{i,t}
=
\mathrm{GRU}(\mathbf{z}_{i,t}, \mathbf{h}_{i,t-1}),
\end{equation}
\begin{equation}
\mathbf{h}_i
=
\mathbf{h}_{i,T_i}
\in \mathbb{R}^{d_h}.
\end{equation}

\textbf{Encoding aggregation and weight prediction.} Once all the encoding completed, all the embedding vectors are combined together to form a final representation of the all the input information as an input for the prediction layer which output the predicted future weights:
\begin{equation}
\tilde{\mathbf{h}}_i
=
\left[
\mathbf{h}_i \, ; \, \mathbf{e}_i^{(c)}
\right].
\end{equation}
\begin{equation}
\mathbf{o}_i
=
\mathbf{W}_2
\, \sigma
\left(
\mathbf{W}_1 \tilde{\mathbf{h}}_i + \mathbf{b}_1
\right)
+
\mathbf{b}_2,
\end{equation}
where 
$\theta(.)$ denotes the ReLU activation.

In our work, we aim to quantify the uncertainty of the prediction, and thus both the mean $\boldsymbol{\mu}_i$ and variance $\boldsymbol{\sigma}_i^2$ of future weights are predicted instead of a single deterministic scaler weight value: 
\begin{equation}
\mathbf{o}_i
=
\begin{bmatrix}
\boldsymbol{\mu}_i \\
\log \boldsymbol{\sigma}_i^2
\end{bmatrix},
\quad
\boldsymbol{\mu}_i, \log \boldsymbol{\sigma}_i^2 \in \mathbb{R}^{H}.
\end{equation}

Accordingly, we select the Gaussian Negative Log-Likelihood (NLL) as the loss function to optimize the prediction model: 
\begin{equation}
\mathcal{L}_{\mathrm{NLL}}
=
\frac{1}{2H}
\sum_{h=1}^{H}
\left[
\log \sigma_{i,h}^2
+
\frac{(y_{i,h} - \mu_{i,h})^2}{\sigma_{i,h}^2}
\right].
\end{equation}
where the predicted future weight $y_{i,h}$ is drawn from the distribution predicted above:
\begin{equation}
y_{i,h}
\sim
\mathcal{N}(\mu_{i,h}, \sigma_{i,h}^2),
\quad h = 1, \dots, H.
\end{equation}

\subsection{Federated Learning Architecture}
We adopt a standard federated averaging (FedAvg) paradigm to train the livestock growth prediction model collaboratively across multiple farms, while keeping all raw livestock production data local. Let $k \in \{1, \dots, K\}$ denote the client (farm) index and
$r \in \{1, \dots, R\}$ the federated communication round. Each client $k$ owns a private dataset $\mathcal{D}_k$ with
$n_k = |\mathcal{D}_k|$ samples, and the total number of samples is
$N = \sum_{k=1}^{K} n_k$. Let $\boldsymbol{\theta}$ denote the full set of model parameters,
including the encoder, GRU layers, and prediction head.

\textbf{Global optimization objective.}
Federated learning implicitly minimizes a data-size–weighted global empirical risk across all clients:
\begin{equation}
\min_{\boldsymbol{\theta}}
\;
\sum_{k=1}^{K}
\frac{n_k}{N}
\;
\mathbb{E}_{(\mathbf{x}, \mathbf{y}) \sim \mathcal{D}_k}
\left[
\ell\!\left(
\mathbf{y},
f(\mathbf{x}; \boldsymbol{\theta})
\right)
\right],
\end{equation} where 
$f(.; \boldsymbol{\theta})$ is the livestock growth prediction model and $\ell(.)$ denotes the local training loss.

\textbf{Server initialization and broadcast.} At the beginning of communication round $r$, the server broadcasts the current global model to all clients:
\begin{equation}
\boldsymbol{\theta}^{(r)}
\;\longrightarrow\;
\text{all clients } k.
\end{equation}

Each client initializes its local model as:
\begin{equation}
\boldsymbol{\theta}_k^{(r,0)} = \boldsymbol{\theta}^{(r)}.
\end{equation}

\textbf{Local model training and update.} Each client performs local optimization for $E$ epochs using its private dataset. For a training sample $i$ at client 
$k$, the model predicts a mean $\mu_{k,i,h})$ and variance $\sigma_{k,i,h}^2$ for each forecast horizon $h$. The local Gaussian negative log-likelihood loss is:
\begin{equation}
\ell_{k,i}
=
\frac{1}{2}
\sum_{h=1}^{H}
\left[
\log \sigma_{k,i,h}^2
+
\frac{(y_{k,i,h} - \mu_{k,i,h})^2}{\sigma_{k,i,h}^2}
\right].
\end{equation}

The empirical local objective at client $k$ is:
\begin{equation}
\mathcal{L}_k(\boldsymbol{\theta})
=
\frac{1}{n_k}
\sum_{i=1}^{n_k}
\ell_{k,i}.
\end{equation}

Starting from the received global parameters, each client performs gradient-based optimization:
\begin{equation}
\boldsymbol{\theta}_k^{(r,E)}
=
\boldsymbol{\theta}^{(r)}
-
\eta
\sum_{e=1}^{E}
\nabla
\mathcal{L}_k
\left(
\boldsymbol{\theta}_k^{(r,e-1)}
\right),
\end{equation}
where $\eta$ denotes the learning rate.
After local training, the updated parameters are denoted as:
\begin{equation}
\boldsymbol{\theta}_k^{(r)} = \boldsymbol{\theta}_k^{(r,E)}.
\end{equation}

\textbf{Federated aggregation (FedAvg) and optimization.} Once all clients complete local training, the server aggregates the received models using data-size–weighted averaging:
\begin{equation}
\boldsymbol{\theta}^{(r+1)}
=
\sum_{k=1}^{K}
\frac{n_k}{N}
\boldsymbol{\theta}_k^{(r)}.
\end{equation}
This aggregation is applied to all layers of the network, ensuring that both temporal dynamics and uncertainty estimates are globally shared. The broadcast–train–aggregate procedure is repeated for $R$ rounds:
\begin{equation}
\boldsymbol{\theta}^{(1)}
\;\rightarrow\;
\boldsymbol{\theta}^{(2)}
\;\rightarrow\;
\cdots
\;\rightarrow\;
\boldsymbol{\theta}^{(R)}.
\end{equation}
The final global model $\boldsymbol{\theta}^{(R)}$
is used for inference on all farms.

\subsection{Personalized Federated Learning Architecture}
To account for farm-specific growth patterns and management practices, we extend the standard federated learning framework with a personalized model head, while keeping the representation learning components globally shared. Specifically, we decompose the model parameters into two parts:
\begin{equation}
\boldsymbol{\theta}
=
\left\{
\boldsymbol{\theta}^{(s)},\;
\boldsymbol{\theta}^{(p)}
\right\},
\end{equation}
where $\theta (s)$ denotes the shared body, including numeric embeddings, GRU layers, and categorical embeddings, $\theta (p)$ denotes the personalized prediction head, implemented as the final fully connected layers.

The federated optimization objective with personalization becomes:
\begin{equation}
\min_{\boldsymbol{\theta}^{(s)}, \{\boldsymbol{\theta}^{(p)}_k\}_{k=1}^{K}}
\;
\sum_{k=1}^{K}
\frac{n_k}{N}
\;
\mathbb{E}_{(\mathbf{x}, \mathbf{y}) \sim \mathcal{D}_k}
\left[
\ell\!\left(
\mathbf{y},
f(\mathbf{x}; \boldsymbol{\theta}^{(s)}, \boldsymbol{\theta}^{(p)}_k)
\right)
\right],
\end{equation}
where each client $k$ maintains its own personalized head $\boldsymbol{\theta}^{(p)}_k$, while sharing the body $\theta (s)$.

In this way, at communication round $r$, the server broadcasts only the shared body:
\begin{equation}
\boldsymbol{\theta}^{(s,r)}
\;\longrightarrow\;
\text{all clients } k.
\end{equation} Each client initializes its local model as:
\begin{equation}
\boldsymbol{\theta}^{(s,r)}_k = \boldsymbol{\theta}^{(s,r)}.
\end{equation} If available, the client restores its previously learned personalized head $\boldsymbol{\theta}^{(p)}_k$.

\textbf{Local training with personalized head.} Each client performs local optimization on its private data, updating both shared and personalized parameters. For sample $i$ at client $k$:
\begin{equation}
\ell_{k,i}
=
\frac{1}{2}
\sum_{h=1}^{H}
\left[
\log \sigma_{k,i,h}^2
+
\frac{(y_{k,i,h} - \mu_{k,i,h})^2}{\sigma_{k,i,h}^2}
\right].
\end{equation}
The empirical local objective is:
\begin{equation}
\mathcal{L}_k
\left(
\boldsymbol{\theta}^{(s)}, \boldsymbol{\theta}^{(p)}_k
\right)
=
\frac{1}{n_k}
\sum_{i=1}^{n_k}
\ell_{k,i}.
\end{equation}

Starting from the broadcast shared body, client $k$ performs gradient-based updates:
\begin{equation}
\boldsymbol{\theta}^{(s,r,E)}_k
=
\boldsymbol{\theta}^{(s,r)}
-
\eta
\sum_{e=1}^{E}
\nabla_{\boldsymbol{\theta}^{(s)}}
\mathcal{L}_k,
\end{equation}
\begin{equation}
\boldsymbol{\theta}^{(p,r,E)}_k
=
\boldsymbol{\theta}^{(p,r)}_k
-
\eta
\sum_{e=1}^{E}
\nabla_{\boldsymbol{\theta}^{(p)}_k}
\mathcal{L}_k.
\end{equation}
After local training, the client retains:
\begin{equation}
\boldsymbol{\theta}^{(p,r)}_k = \boldsymbol{\theta}^{(p,r,E)}_k,
\end{equation}
while sending only the shared parameters $\boldsymbol{\theta}^{(s,r)} =\boldsymbol{\theta}^{(s,r,E)}_k$ to the server.

The server aggregates only the shared body parameters using FedAvg:
\begin{equation}
\boldsymbol{\theta}^{(s,r+1)}
=
\sum_{k=1}^{K}
\frac{n_k}{N}
\boldsymbol{\theta}^{(s,r)}_k.
\end{equation}
The personalized heads $\boldsymbol{\theta}^{(p)}_k$ are excluded from aggregation and remain private to each client.

The personalized federated learning process repeats for 
$R$ rounds:
\begin{equation}
\boldsymbol{\theta}^{(s,1)}
\;\rightarrow\;
\boldsymbol{\theta}^{(s,2)}
\;\rightarrow\;
\cdots
\;\rightarrow\;
\boldsymbol{\theta}^{(s,R)}.
\end{equation}
At convergence, the final model for client $k$ is:
\begin{equation}
f_k(\mathbf{x})
=
f(\mathbf{x}; \boldsymbol{\theta}^{(s,R)}, \boldsymbol{\theta}^{(p)}_k).
\end{equation}

For inference on farm 
$k$, predictions are generated using the shared body and the farm-specific head:
\begin{equation}
\hat{\mathbf{y}}_k
=
f(\mathbf{x}; \boldsymbol{\theta}^{(s,R)}, \boldsymbol{\theta}^{(p)}_k).
\end{equation}

\vspace{1pt}
\section{Experiments}

\subsection{Dataset}

We conducted intensive experiments on a real-world livestock production dataset collected in 
Australia in collaboration with a local livestock management service provider. After processing, the dataset contains 25,422 beef cattle from 110 farms distributed across 10 NRM regions in New South Wales. The cattle are primarily of the Angus breed and its crosses, including ’Angus’, ’Angus X’, ’Angus Hereford X’, ’Red Angus’, ’Angus Friesian X’, ’South Devon Angus X’, ’Angus Lowline’, ’Brahman Angus X’, and ’Australis South Devon/Angus’. Four categorical static livestock features are included: sex, breed, state, and NRM region, with more features (e.g., climate) easily incorporable when available.

For sequence data, we use historical weight records from each animal between 2 and 24 months of age, the main growth stage of beef cattle. In the raw dataset, most animals have very sparse weight records, averaging only 2.5 measurements per animal. To address this, we designed a novel quantile regression-based inter- and extrapolation method to generate monthly weight records from 2 to 24 months. To improve reliability, an age-based distance is calculated between each augmented record and its nearest observed weight record, and a distance-based credibility is assigned: smaller distances correspond to higher credibility. Min-max normalization is applied to weight, age, distance, and credibility, and a binary masking vector indicates whether a weight record is observed at each time step.

The final dataset statistics are as follows: 20,337 animals contribute to 77,987 training instances, and 5,085 animals contribute to 19,502 testing instances.

\subsection{Experiment Settings}
We aim to answer the following research questions (RQ) through experiments: 

RQ 1: Comparison of prediction accuracy: How is our proposed method compared to centralized training? How does our personalized federated learning perform?  


RQ 2: How our proposed model can effectively alleviates data
sparsity and data scarcity issues, particularly for small farms (which is a typical issue in our current project)? 

RQ 3: Who benefits most from our propose method, small farms or large farms? (Small vs large farm analysis) and how our method can benefit farms of different sizes differently?

\subsection{Experiment Results and Analysis}

\subsubsection{Weight prediction accuracy comparison under different settings (RQ1)}: We evaluate the GRU-based livestock growth prediction model under three settings to understand its overall performance: (1) Centralized training, serving as the reference baseline; (2) Federated learning (LivestockFL), where each client’s contribution to global model updates is weighted by its sample size; and (3) Personalized federated learning (LivestockPFL), an enhanced version of LivestockFL designed to address imbalanced and heterogeneous data distributions across farms.

To further mitigate the impact of highly imbalanced client sizes, we also introduce LivestockFL-Sqrt and LivestockPFL-Sqrt, variants that apply $\sqrt{n}$ weighting in FedAvg. Specifically, instead of weighting each client’s model by its sample  size $n_i$, the weight is proportional to the square root of the sample size:
$w_i = \frac{\sqrt{n_i}}{\sum_{j=1}^{K} \sqrt{n_j}}$. This $\sqrt{n}$ weighting reduces the dominance of large clients in standard federated learning and helps alleviate potential bias in the aggregated prediction models.

\begin{table}[b]
\caption{Overall comparisons on livestock weight prediction accuracy on the whole dataset}
\centering
\begin{tabular}{|
    c
    |p{1cm}
    |p{1cm}
    |p{1cm}
    |p{1cm}|
}
\hline
\cline{2-5}
 & \textbf{\textit{RMSE}}
 & \textbf{\textit{MAE}}
 & \textbf{\textit{MAPE}}
 & \textbf{\textit{R-score}} \\
\hline
 \textbf{\textit{Centralized training}}   & 14.34 &10.30 &2.71\% &0.97 \\
\hline
\hline
\textbf{\textit{LivestockFL}}   & 42.76 &33.46
 &9.13\% & 0.75\\ 
\hline
\textbf{\textit{LivestockFL-Sqrt}}   & 39.55 &31.84
 &8.53\% & 0.80\\ 
 \hline
\textbf{\textit{LivestockPFL}}   & \textbf{23.42}
 &\textbf{17.33} &\textbf{4.66\%} & \textbf{0.93}\\
\hline
\textbf{\textit{LivestockPFL-Sqrt}}   & 26.81
 &20.41 &5.42\% & 0.91
\\  
\hline
\end{tabular}
\label{overall_accuracy}
\end{table}

\begin{table}[htbp]
\caption{Per-prediction horizon comparisons on livestock weight prediction accuracy on the whole dataset}
\centering
\begin{tabular}{|
    p{1.1cm}
    |p{1.7cm}
    |p{0.8cm}
    |p{0.8cm}
    |p{0.8cm}
    |p{0.9cm}|
}
\hline
\cline{3-6}
\textbf{Prediction Horizon} & \textbf{Method}
 & \textbf{\textit{RMSE}}
 & \textbf{\textit{MAE}}
 & \textbf{\textit{MAPE}}
 & \textbf{\textit{R-score}} \\
\hline

\multirow{3}{*}{\textbf{Horizon 1}}
 & \textbf{\textit{Centralized training}} & 9.24 &6.77 &1.90\% &0.98 \\
\hhline{|~=====|}
 & \textbf{\textit{LivestockFL}} & 46.22  &36.55  &10.44  & 0.72 \\
\cline{2-6}
 & \textbf{\textit{LivestockFL-Sqrt}} & 39.49  &32.29  &9.16  & 0.80 \\
  
\cline{2-6}
 & \textbf{\textit{LivestockPFL}} & \textbf{19.71} & \textbf{14.78} & \textbf{4.17\%} & \textbf{0.95} \\
\cline{2-6}
 & \textbf{\textit{LivestockPFL-Sqrt}} & 24.43 & 19.20 & 5.32\% &  0.92 \\
\hline

\multirow{3}{*}{\textbf{Horizon 2}}
 & \textbf{\textit{Centralized training}} & 14.08 & 10.39 & 2.77\% & 0.97 \\
\hhline{|~=====|}
 & \textbf{\textit{LivestockFL}} & 42.01 & 32.65 & 8.84\% & 0.77 \\
\cline{2-6}
 & \textbf{\textit{LivestockFL-Sqrt}} & 40.07 & 31.61 & 8.50\% & 0.79 \\
\cline{2-6}
 & \textbf{\textit{LivestockPFL}} &  \textbf{23.03} & \textbf{16.88} & \textbf{4.55\%} & \textbf{0.93} \\
\cline{2-6}
 & \textbf{\textit{LivestockPFL-Sqrt}} & 25.88 & 19.51 & 5.18\% & 0.91 \\
\hline

\multirow{3}{*}{\textbf{Horizon 3}}
 & \textbf{\textit{Centralized training}} & 18.25 & 13.74 & 3.45\% & 0.95 \\
\hhline{|~=====|}
 & \textbf{\textit{LivestockFL}} & 39.79 & 31.18 & 8.08\% & 0.78 \\
\cline{2-6}
 & \textbf{\textit{LivestockFL-Sqrt}} & 39.08 & 31.63 & 7.95\% & 0.79 \\
\cline{2-6}
 & \textbf{\textit{LivestockPFL}} & \textbf{26.96} & \textbf{20.33} & \textbf{5.26\%} &  \textbf{0.90} \\
\cline{2-6}
 & \textbf{\textit{LivestockPFL-Sqrt}} & 29.82 & 22.51 & 5.76\% & 0.88 \\
\hline

\end{tabular}
\label{horizon_accuracy}
\end{table}

\textbf{Overall prediction performance (Table \ref{overall_accuracy})}: Table I shows the overall weight prediction accuracy across all animals and time horizons. Centralized training achieves the best performance, with RMSE = 14.34kg, MAE = 10.30kg, MAPE = 2.71\%, and $R^2$ = 0.97, as expected since it has full access to all data. Standard federated learning (LivestockFL) performs substantially worse (RMSE = 42.76kg, MAE = 33.46kg, MAPE = 9.13\%, $R^2$ = 0.75), reflecting the challenges of imbalanced and heterogeneous farm-level data across the 110 farms, where small farms’ contributions are underrepresented. Introducing the $\sqrt{n}$ weighting (LivestockFL-Sqrt) improves FL performance (RMSE = 39.55kg, MAE = 31.84kg, MAPE = 8.53\%, $R²$ = 0.80), confirming that down-weighting the dominance of very large farms mitigates aggregation bias.

Personalized FL (LivestockPFL) significantly improves accuracy over standard FL, achieving RMSE = 23.42kg, MAE = 17.33kg, MAPE = 4.66\%, $R²$ = 0.93. Its $\sqrt{n}$ variant (LivestockPFL-Sqrt) also performs well, although slightly lower than standard PFL (RMSE = 26.81kg, MAE = 20.41kg, MAPE = 5.42\%, R² = 0.91). This demonstrates that personalization is highly effective in heterogeneous and imbalanced data scenarios, allowing the model to adapt to farm-specific characteristics while preserving privacy. Overall, LivestockPFL bridges a large portion of the gap between FL and centralized training.

\textbf{Per-horizon prediction performance (Table \ref{horizon_accuracy})}: Table II further reveals trends across three prediction horizons. Centralized training consistently outperforms all FL methods for each horizon, with RMSE increasing gradually from horizon 1 (9.24kg) to horizon 3 (18.25kg), reflecting the increased difficulty in longer-term predictions. Standard FL (LivestockFL) shows the largest errors in horizon 1 (RMSE = 46.22kg) but improves slightly across longer horizons, likely due to averaging effects. The $\sqrt{n}$ variant (LivestockFL-Sqrt) reduces horizon errors noticeably, especially for horizon 1 (RMSE = 39.49kg), validating its bias-mitigation effect.

Personalized FL methods (LivestockPFL and LivestockPFL-Sqrt) achieve consistently lower errors across all horizons, with PFL achieving RMSE = 19.71kg, 23.03kg, and 26.96kg for horizons 1, 2, and 3, respectively. The improvement over standard FL is particularly pronounced in the early horizon, demonstrating the ability of PFL to leverage farm-specific information for more accurate short-term predictions, while maintaining robustness over longer horizons.

\subsubsection{Prediction error analysis on small farms with extreme data scarcity (for RQ2)} 
\begin{figure}[htbp]
\centering{\includegraphics[width=0.5\textwidth]{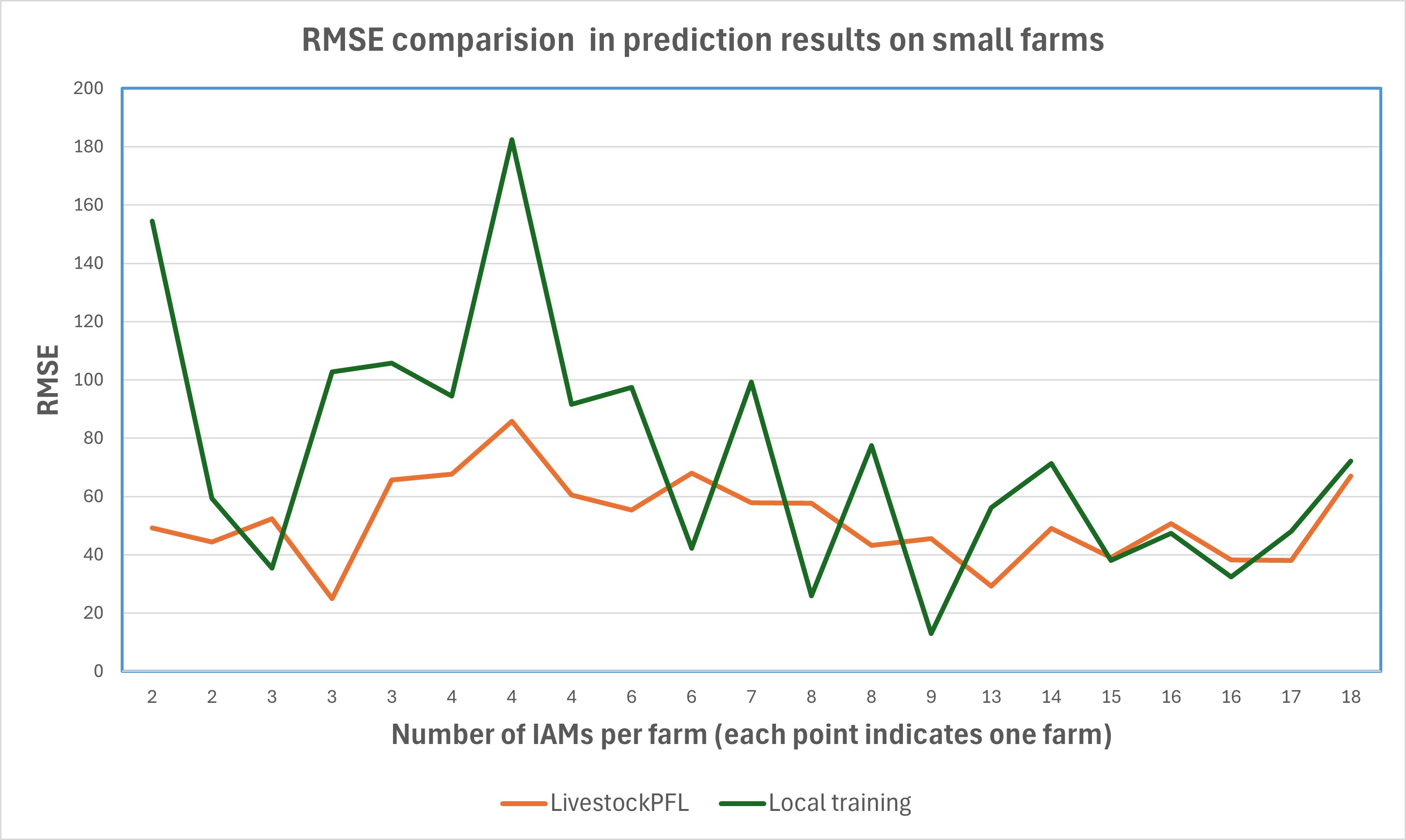}}
\caption{RMSE comparison on small farms.}
\label{RMSE_compare}
\end{figure}

\begin{figure}[htbp]
\centering{\includegraphics[width=0.5\textwidth]{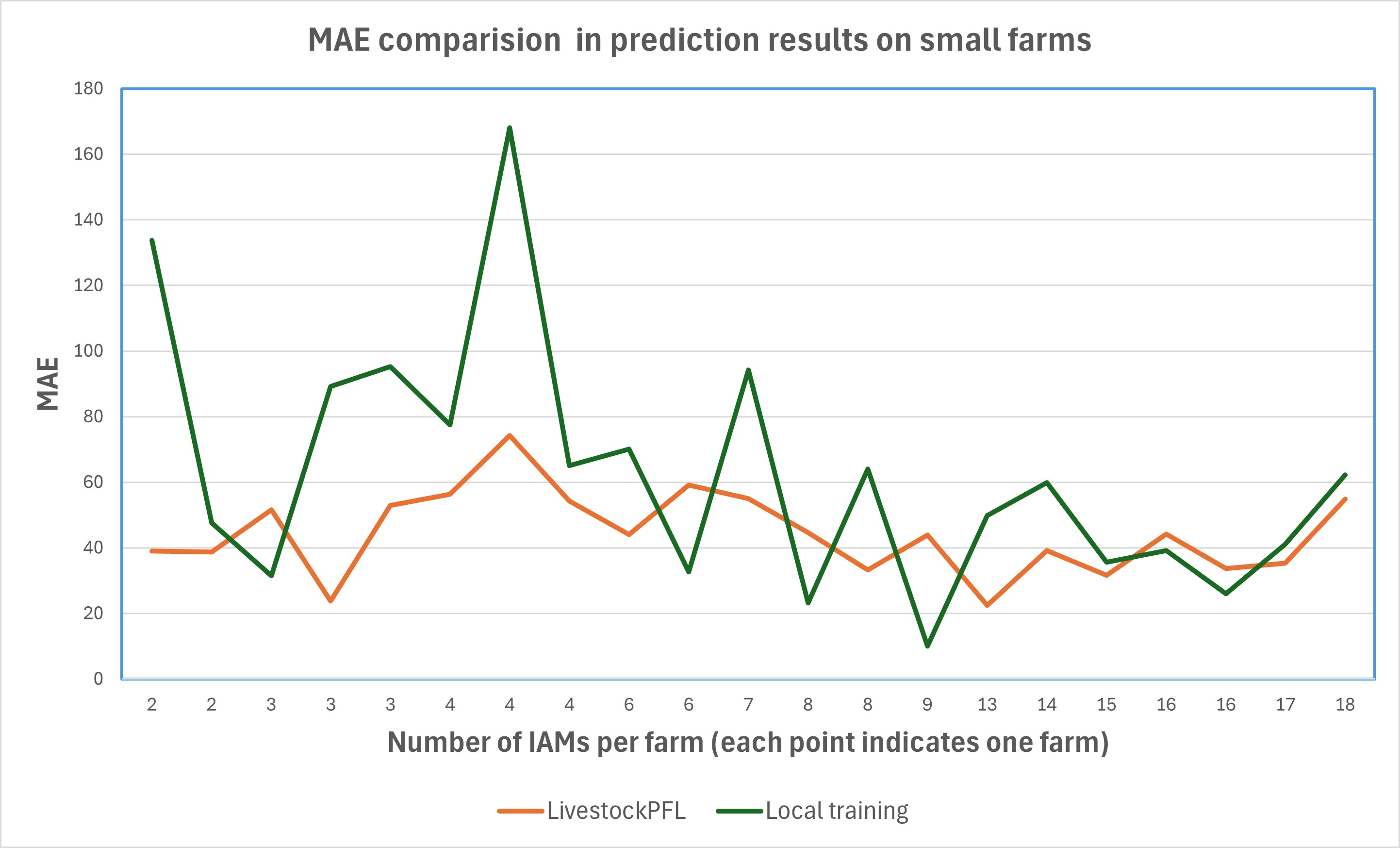}}
\caption{MAE comparison on small farms.}
\label{MAE_compare}
\end{figure}

We evaluated the performance of LivestockPFL against local training on small farms characterized by extreme data scarcity ($<20$ individual animal measurements (IAMs)). As illustrated in the Figs \ref{RMSE_compare} and \ref{MAE_compare}, LivestockPFL consistently outperforms local training in both predictive accuracy and model stability.
Local training exhibits high volatility, with error metrics spiking significantly ($RMSE \approx180$, $MAE \approx 170$) at low IAM counts of 4. This instability underscores the failure of local models to generalize when trained on sparse, high-variance datasets. In contrast, LivestockPFL maintains a substantially lower and more consistent error profile across the entire spectrum of data scarcity. By leveraging the federated framework to capture global weight-gain patterns while personalizing the model to local farm conditions, our approach effectively mitigates the ``cold-start" problem.
The results demonstrate that LivestockPFL provides a robust ``knowledge floor," preventing the performance collapse typical of local deep learning models in data-poor environments. We conclude that the proposed personalized federated learning architecture successfully alleviates data sparsity issues, making accurate livestock weight prediction accessible and reliable for small-scale enterprises that lack sufficient local data for independent model training.

\subsubsection{Small vs. large farm analysis (For RQ3)}
Table \ref{tab:farm_size_comparison_improve} reports performance comparisons between the proposed personalised federated learning method LivestockPF and local training across five farm-size categories defined by the number of IAMs. A clear and consistent pattern emerges: the benefit of the proposed method strongly depends on farm size.

For small farms ($<50$ IAMs), PFL improves performance on 4 out of 5 farms (80\%). In this regime, local models often exhibit high prediction error and unstable or even negative $R^2$ values, reflecting severe data scarcity. By contrast, PFL substantially reduces RMSE, MAE, and MAPE while yielding consistently positive $R^2$, indicating improved generalisation through cross-farm knowledge transfer. For 51–200 IAMs and 201–500 IAMs, the proposed method achieves improvements on all farms (100\% improvement rate in both groups). Error reductions are consistent across RMSE, MAE, and MAPE, and $R^2$ values are uniformly high. These results suggest that medium-sized farms represent the most favourable regime for PFL, where sufficient local data enables effective personalisation while still benefiting from shared global representations.

In contrast, the advantage diminishes for 501–1000 IAMs, where only 1 out of 5 farms (20\%) shows improvement. For most farms in this group, local training already achieves strong performance, limiting the marginal gains from federated learning. For large farms ($>1000$ IAMs), no improvements are observed. Local models consistently outperform or match PFL across all metrics, indicating that abundant local data largely eliminates the need for external knowledge transfer.

Overall, these results demonstrate that the proposed method benefits small and medium farms most, while remaining competitive but unnecessary for very large, data-rich farms.

\begin{table*}[htbp]
\centering
\caption{Performance comparison between Personalised Federated Learning (PFL) and Local Training across different farm-size levels. IAMs refer to individual animals. The last column indicates whether PFL improves over local training (1 = yes, 0 = no).}
\label{tab:farm_size_comparison_improve}
\renewcommand{\arraystretch}{1.1}
\setlength{\tabcolsep}{4.5pt}
\begin{tabular}{|l|c|cccc|cccc|c|}
\hline
\multirow{2}{*}{Farm-size level} & \multirow{2}{*}{\# IAMs}
& \multicolumn{4}{c|}{Personalised FL}
& \multicolumn{4}{c|}{Local Training}
& \multirow{2}{*}{Improve} \\
\cline{3-10}
 &  & RMSE & MAE & MAPE(\%) & $R^2$
 & RMSE & MAE & MAPE(\%) & $R^2$ &  \\
\Xhline{1.2pt}

\multirow{5}{*}{\textless 50 IAMs}
 & 6  & 55.35 & 44.11 & 11.84 & 0.59 & 97.46 & 70.16 & 20.51 & -0.26 & 1 \\ \cline{2-11}
 & 14 & 49.00 & 39.16 & 11.13 & 0.60 & 71.27 & 59.96 & 17.55 &  0.15 & 1 \\ \cline{2-11}
 & 26 & 53.38 & 46.83 & 14.57 & 0.13 & 51.69 & 40.72 & 11.65 &  0.18 & 0 \\ \cline{2-11}
 & 38 & 29.46 & 22.31 & 10.44 & 0.85 & 38.60 & 29.55 & 12.23 &  0.75 & 1 \\ \cline{2-11}
 & 46 & 24.69 & 19.92 &  6.42 & 0.93 & 51.26 & 39.00 & 13.54 &  0.69 & 1 \\
\Xhline{1.2pt}

\multirow{5}{*}{51--200 IAMs}
 & 55  & 37.82 & 29.12 &  6.94 & 0.42 & 56.83 & 46.44 & 11.02 & -0.31 & 1 \\ \cline{2-11}
 & 70  & 48.98 & 37.21 &  8.63 & 0.26 & 61.30 & 46.26 & 11.77 & -0.15 & 1 \\ \cline{2-11}
 & 92  & 24.53 & 18.97 &  5.34 & 0.91 & 30.00 & 24.62 &  7.05 &  0.86 & 1 \\ \cline{2-11}
 & 127 & 22.97 & 17.75 &  4.19 & 0.83 & 27.98 & 24.50 &  5.61 &  0.75 & 1 \\ \cline{2-11}
 & 174 & 19.07 & 14.91 &  5.23 & 0.95 & 19.24 & 14.80 &  4.97 &  0.95 & 1 \\
\Xhline{1.2pt}

\multirow{5}{*}{201--500 IAMs}
 & 214 & 25.56 & 19.43 &  5.92 & 0.93 & 30.63 & 24.14 &  6.55 & 0.90 & 1 \\ \cline{2-11}
 & 245 & 14.97 & 11.98 &  4.37 & 0.96 & 25.63 & 19.85 &  6.64 & 0.89 & 1 \\ \cline{2-11}
 & 361 & 17.91 & 12.94 &  3.72 & 0.96 & 22.44 & 17.51 &  5.12 & 0.94 & 1 \\ \cline{2-11}
 & 408 & 20.14 & 16.77 &  4.50 & 0.94 & 21.13 & 17.49 &  4.75 & 0.93 & 1 \\ \cline{2-11}
 & 490 & 23.59 & 18.34 &  4.36 & 0.89 & 59.26 & 55.30 & 13.58 & 0.32 & 1 \\
\Xhline{1.2pt}

\multirow{5}{*}{501--1000 IAMs}
 & 502 & 11.94 &  9.71 &  2.84 & 0.97 & 10.21 &  7.91 &  2.34 & 0.98 & 0 \\ \cline{2-11}
 & 747 & 20.62 & 15.90 &  5.05 & 0.95 & 23.65 & 17.94 &  5.51 & 0.93 & 1 \\ \cline{2-11}
 & 837 & 16.14 & 12.96 &  3.16 & 0.96 & 16.02 & 12.90 &  3.20 & 0.96 & 0 \\ \cline{2-11}
 & 883 & 43.33 & 38.99 & 12.08 & 0.75 & 28.40 & 23.15 &  7.41 & 0.89 & 0 \\ \cline{2-11}
 & 969 & 25.16 & 21.18 &  5.23 & 0.93 & 17.84 & 14.20 &  3.54 & 0.96 & 0 \\
\Xhline{1.2pt}

\multirow{5}{*}{\textgreater 1000 IAMs}
 & 1796 & 18.07 & 13.74 & 3.78 & 0.93 & 16.98 & 13.49 & 3.83 & 0.93 & 0 \\ \cline{2-11}
 & 2004 & 22.60 & 17.94 & 4.34 & 0.90 & 22.04 & 16.91 & 4.00 & 0.90 & 0 \\ \cline{2-11}
 & 2667 & 19.24 & 14.89 & 3.94 & 0.94 & 15.90 & 11.87 & 3.16 & 0.96 & 0 \\ \cline{2-11}
 & 3279 & 23.80 & 18.02 & 4.42 & 0.91 & 16.88 & 12.67 & 3.05 & 0.96 & 0 \\ \cline{2-11}
 & 3419 & 23.39 & 16.90 & 3.94 & 0.93 & 13.37 &  9.58 & 2.31 & 0.98 & 0 \\
\hline
\end{tabular}
\end{table*}


\section{Conclusions}

Livestock growth prediction is essential for optimizing farm management and improving both efficiency and sustainability. In practice, livestock production data are often fragmented and distributed across multiple farms. Effectively leveraging this data for reliable growth prediction while addressing data privacy and security concerns remains a significant challenge, with limited studies addressing it. In this paper, we propose a novel neural federated learning framework as a promising solution. To further tackle data imbalance and heterogeneity, we extend this framework to a personalized federated learning approach. To the best of our knowledge, this is the first public study applying federated learning to livestock growth prediction. Extensive experiments on a real-world livestock production dataset demonstrate the effectiveness of our proposed methods. Future work will explore more advanced federated learning models for livestock growth prediction, and collect and investigate more real-world livestock datasets.
\bibliographystyle{IEEEbib}
\bibliography{Ref}
\end{document}